%% file: main.tex
\pdfoutput=1

\documentclass[11pt]{article}

\usepackage[preprint]{acl}

\usepackage{times}
\usepackage{latexsym}
\usepackage{graphicx}
\usepackage{fancyvrb}
\usepackage{xcolor}
\usepackage[T1]{fontenc}

\usepackage[utf8]{inputenc}

\usepackage{microtype}

\usepackage{inconsolata}
\usepackage{amsmath}
\usepackage{amssymb}
\usepackage{authblk}

\title{$\mathbf{(N,K)}$-Puzzle: A Cost-Efficient Testbed for Benchmarking Reinforcement Learning Algorithms in Generative Language Models}

\author[1]{Yufeng Zhang}
\author[1]{Liyu Chen}
\author[1]{Boyi Liu}
\author[1]{Yingxiang Yang}
\author[2]{Qiwen Cui\thanks{Work is done while at ByteDance.}}
\author[1]{Yunzhe Tao}
\author[1]{Hongxia Yang}

\affil[1]{ByteDance, Inc. }
\affil[2]{University of Washington, Seattle
\authorcr \{yufeng.zhang,liyu.chen1,boyi.liu01,yingxiang.yang,yunzhe.tao,hx.yang\}@bytedance.com
\authorcr qwcui@uw.edu }

\begin{document}
\maketitle
\begin{abstract}
Recent advances in reinforcement learning (RL) algorithms aim to enhance the performance of language models at scale. Yet, there is a noticeable absence of a cost-effective and standardized testbed tailored to evaluating and comparing these algorithms.
To bridge this gap, we present a generalized version of the 24-Puzzle: the $(N,K)$-Puzzle, which challenges language models to reach a target value $K$ with $N$ integers.
We evaluate the effectiveness of established RL algorithms such as Proximal Policy Optimization (PPO), alongside novel approaches like Identity Policy Optimization (IPO) and Direct Policy Optimization (DPO).
\end{abstract}

\section{Introduction}

Reinforcement learning (RL) is an important component in the training process of language models (LMs) \citep{achiam2023gpt, team2023gemini, touvron2023llama, touvron2023llama2, bi2024deepseek, shao2024deepseekmath}. Integrated into the training process subsequent to the supervised fine-tuning (SFT) phase,
RL significantly enhances generative LMs' performance by incorporating feedback from sources such as human labelers \citep{stiennon2020learning, ouyang2022training}, other generative LMs \citep{lee2023rlaif}, or automated test environments \citep{le2022coderl}. However, despite its importance, there lacks a benchmark that is targeted at isolating and testing the performance of the RL phase of the training {\it alone}. As a matter of fact, the prevailing spectrum of benchmarks that evaluate the LMs capabilities \citep{hendrycks2020measuring, hendrycks2021measuring, chen2021evaluating, cobbe2021training} contain complex and multifaceted influencing factors ranging from the capability of the pretrained models and the SFT models to the datasets and algorithms. These factors prevents analyzing the impact of RL in an isolated fashion, and a cost-effective and standardized testbed for the comparison of different RL methods applied to generative LMs is still arguably missing.


In this paper, we introduce the $(N, K)$-Puzzle, a generalized version of the classic 24-Puzzle, as an efficient and streamlined testbed to evaluate RL strategies in training generative LMs. Our framework offers flexibility through adjustable parameters $N$ and $K$, making it capable to systematically explore the efficacy, performance, and scalability of RL strategies. Additionally, manipulating $N$ and $K$ enables a thorough assessment of the generalization abilities of different RL methods, which provides valuable insights into their potential applicability and effectiveness in more complex tasks.

\section{Background}
\subsection{Reward Model Training}

We train a reward model (RM) from a preference dataset $\mathcal{D}_{\text{pref}} = \{(x^{(i)}, y_w^{(i)}, y_l^{(i)})\}_{i = 1}^N$, where $x^{(i)}$ is the prompt, $y_w^{(i)}$ is the chosen (win) response, and $y_l^{(i)}$ is the rejected (lose) responses. We parameterize the RM as $r_\phi(x, y)$ and learn the parameters $\phi$ by minimizing the negative log-likelihood loss
\begin{align*}
    - \underset{(x, y_w, y_l) \sim \mathcal{D}_{\text{pref}}}{\mathbb{E}} \Bigl[\log \sigma\bigl( r_\phi(x, y_w) - r_\phi(x, y_l) \bigr)\Bigr],
\end{align*}
where $\sigma$ is the sigmoid function.

\subsection{\bf Reinforcement Learning with an RM}

The RL phase trains on a dataset $\mathcal{D} = \{x^{(i)}\}$ where $x^{(i)}$ is a prompt. We denote the LM policy as $\pi_\theta$ and optimize $\theta$ through a KL regularized objective:
\begin{align}
    \label{eq:rl}
    \underset{x\sim \mathcal{D}, y \sim \pi_\theta(\cdot | x)}{\mathbb{E}}\bigl[r_\phi(x, y)\bigr] - \beta \cdot \mathbb{D}_{\mathrm{KL}} ( \pi_\theta \| \pi_{\mathrm{ref}} ),
\end{align}
where $\beta > 0$ controls the deviance between the trained model $\pi_\theta$ and the reference model $\pi_{\mathrm{ref}}$. A standard approach \citep{ouyang2022training,ziegler2019fine} to solve \eqref{eq:rl} is to initialize the LM policy $\pi_\theta$ to the SFT model $\pi^{\mathrm{SFT}}$ and maximize the objective with PPO \citep{schulman2017proximal}.

\subsection{Reinforcement Learning without an RM}

Recently developed RL algorithms, such as direct preference optimization (DPO) \citep{rafailov2023direct} and identity preference optimization (IPO) \citep{azar2023general}, optimize the policy directly from the preference dataset without learning a reward model. Particularly, DPO combines the training of the RM and the reinforcement learning by minimizing a unified loss
\begin{align}
    \label{eq:dpo}
      \underset{(x, y_w, y_l) \sim \mathcal{D}_{\text{pref}}}{\mathbb{E}} \Bigl[\log \sigma\bigl(\beta \cdot h_{\pi_\theta, \pi_{\mathrm{ref}}}(y_w, y_l, x)\bigr) \Bigr],
\end{align}
where 
\begin{align*}
    h_{\pi_\theta, \pi_{\mathrm{ref}}}(y_w, y_l, x) = \log\frac{\pi_\theta(y_w | x)\pi_{\mathrm{ref}}(y_l | x)}{\pi_\theta(y_l | x) \pi_{\mathrm{ref}}(y_w | x)}.
\end{align*}
Alternatively, \citet{azar2023general} show that DPO suffers from weak regularization and overfitting and generalize the DPO objective to $\Psi$-preference optimization ($\Psi$PO). By setting $\Psi$ to identity, \citet{azar2023general} propose IPO, which optimizes $\theta$ to minimize the following loss,
\begin{align}
    \label{eq:ipo}
    \underset{(x, y_w, y_l) \sim \mathcal{D}_{\text{pref}}}{\mathbb{E}} \Bigl[\Bigl(h_{\pi_\theta, \pi_{\mathrm{ref}}}(y_w, y_l, x) - \frac{1}{2\beta} \Bigr)^2 \Bigr],
\end{align}
where $\beta > 0$ is a regularization parameter.

In the remainder of the paper, we test and compare these aforementioned algorithms on the $(N,K)$-Puzzle, which we now introduce.
\section{Problem Setup: $\mathbf{(N, K)}$-Puzzle}
The $(N,K)$-Puzzle generalizes the classic 24-Puzzle, where arithmetic operations ($+$, $-$, $*$, $/$) are utilized to manipulate a set of \( N \) numbers, $\{x_1,\ldots, x_N\}$, to achieve a target value \( K \). On top of generating the correct answer, the generative LM is also expected to output a sequence of arithmetic operations is mathematically correct. Aligning with the classic 24-Puzzle setting, we restrict $x_i$ to between 1 and 13. An illustrative example of the prompt and the response for the \((4, 24)\)-Puzzle problem is presented below.
\begin{description}
    \item {\it Prompt:} 24;2,3,4,6:
    \item {\it Response:} 4+6=10,10-2=8,8*3=24
\end{description}
This setup challenges the model's computational abilities, its capacity for logical reasoning, and its ability to generate coherent, step-by-step solutions to mathematically grounded problems. In our study, we explore the model's performance across a range of scenarios where the number of operands \( N \) varies from 3 to 4, and the target value \( K \) is chosen from the set $\{13, 18, 24, 27, 34\}$. This investigation is designed to assess the model's adaptability and robustness in handling variations in problem complexity and objectives.

We conduct evaluations in two distinct test set: in-distribution test set and out-of-distribution (OOD) test set. In particular, for the in-distribution test set, while the prompts are not presented in the training set, the pair $(N, K)$ that defines the puzzle is presented to the model in training. For the OOD test set, the pair $(N, K)$ is not presented to the model in training. These two test sets allow us to comprehensively assess the model's capability from both the in-distribution generalization and the OOD generalization. In particular, we assess the difficulty of each $(N, K)$-Puzzle and choose $(3, 18)$-Puzzle and $(4, 27)$-Puzzle for the OOD test set, which are at the medium difficulty among all $(N, K)$-Puzzles.
    

\section{Experiments}
\begin{figure}
    \centering
    \includegraphics[width=\linewidth]{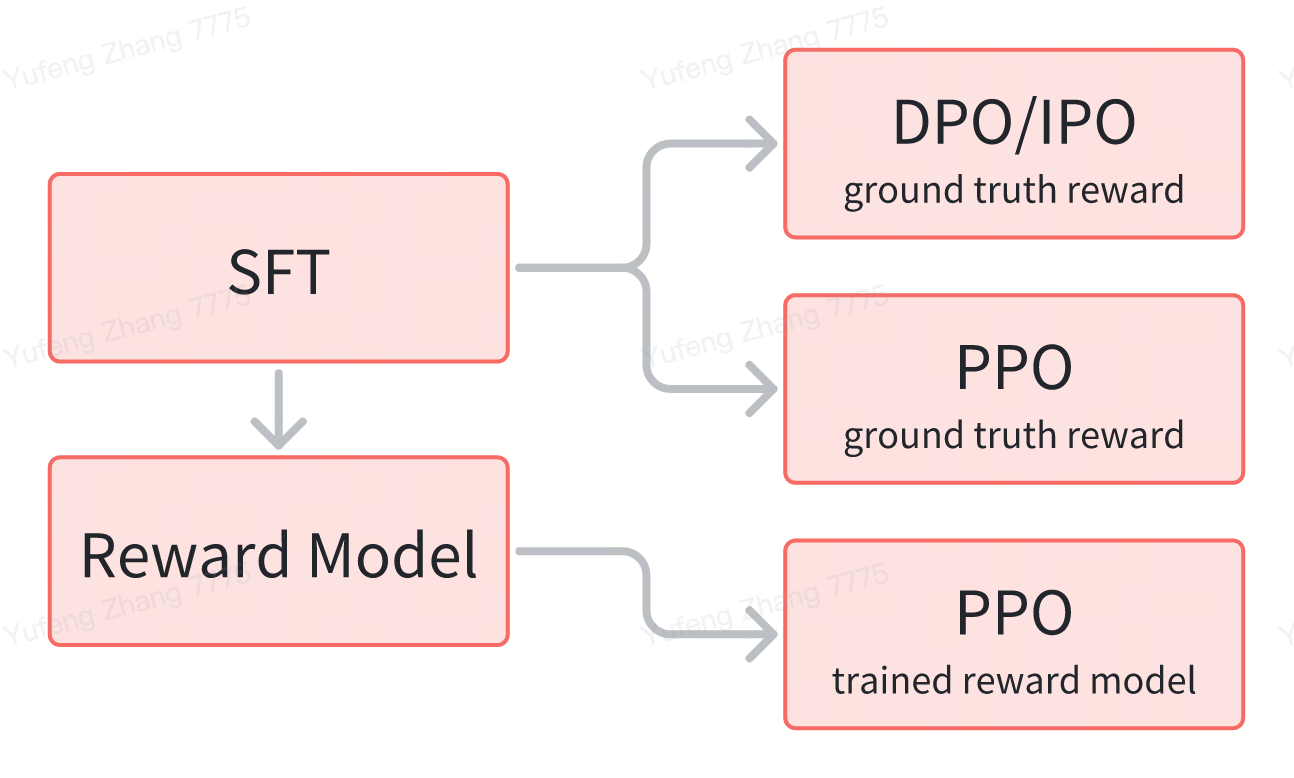}
    \caption{Training of the model: The process of training consists of multiple parts. Firstly, we train the model with SFT to align the model with the $(N, K)$-Puzzle. Then, we perform RL training (PPO, DPO, IPO, and RM) based on the SFT model.}
    \label{fig:pipe}
\end{figure}

\subsection{Experiment Setup}

\noindent{\bf Model.} We utilize the GPT-2 \citep{radford2019language} architecture. The specific instantiation of the GPT-2 model employed in our study is defined by the following parameters: the model comprises \(n_{\text{layer}} = 12\) transformer layers, with each layer featuring \(n_{\text{head}} = 12\) attention heads. The dimensionality of the embeddings and the hidden layers is set to \(n_{\text{embd}} = 768\). This results in a total of approximately 124 million trainable parameters.

\noindent{\bf Supervised fine-tuning.} 
The initial phase of our training is the supervised fine-tuning (SFT). We conduct two stages of SFT from scratch: format SFT and target SFT. In format SFT, we  train the model from scratch using specially formatted data to inculcate the ability to output results in a predefined format and to execute arithmetic operations. Note that during the format SFT phase, the model's outputs may not always align perfectly with the target results. In the format SFT phase, we employ a training dataset consisting of 200,000 data points, encompassing a broad spectrum of arithmetic operations. The model is trained for 20 epochs with a global batch size of 256. We use a learning rate of $10^{-5}$ and a constant learning rate scheduler. Upon completion of this phase, the model demonstrates a remarkable capability to generate responses that outputs responses with correct arithmetic operations and desired formats an accuracy rate of 99\%.

Building on the format SFT phase, we train the model further through target SFT. Initiated from the checkpoint from the format SFT, the objective is to hone the model's capabilities not only to execute arithmetic operations within the specified format accurately but also to engage in logical reasoning processes that ensure the final result aligns with the target value \( K \). During the target SFT phase, we employ a training dataset consisting of 300,000 data points. The model is trained for 10 epochs with a global batch size of 128. We use a learning rate of $10^{-5}$ and a constant learning rate scheduler. Upon completion of this phase, the model achieves an accuracy of 43.5\% in the in-distribution test set and an accuracy of 8.8\% in the OOD test set.

\noindent{\bf Ground truth reward.} 
We define the ground truth reward function \( r^*(x, y) \) to evaluate the responses generated by the model as follows,
\begin{align*}
    r^*(x, y) = \begin{cases}
        1.0,  \text{~ if } y \text{ is completely correct},\\
        0.1, \text{~ if } y \text{ is correct in format}, \\
        0.0, \text{~ otherwise}.
    \end{cases}
\end{align*}
To further illustrate the mechanism of the ground truth reward function, we include examples of responses with different ground truth rewards in \S\ref{sec:eg-reward}.
This ground truth reward function \( r^*(x, y) \) is employed to label the preference dataset for training RM, DPO, and IPO. Furthermore, it is utilized in experiments involving PPO to establish a performance upper bound for PPO with RM.


\begin{table}[]
    \centering
    \begin{tabular}{c|c|c|c|c}
    \hline
          & training & test & OOD & OOD (n=20) \\
         \hline
         RM & $62.5$ & $60$ & $10.5$ & $12.7$\\
         \hline
         GT & $64.6$  & $61.9$ & $14.4$ & $26.9$ \\
         \hline
    \end{tabular}
    \caption{Accuracy (in percentage) of Best-of-$n$ on training dataset ($n=5$), in-distribution test set ($n=5$), and OOD test set ($n=5,20$).}
    \label{tab:my_label}
\end{table}
\subsection{Reward Model}
The training data of our RM is generated as follows: for each prompt in the training set of the SFT model, we collect 10 unique responses from the SFT model, along with their ground truth rewards. If all these 10 responses have the same reward, we sample another 10 responses to make sure that the collected responses always provide training signals. We then train a RM using this training set with batch size $128$ for $500$ iterations. Each batch contains $128$ prompts, and we group all responses for each prompt into the same batch. To test the accuracy of the RM, we perform best-of-$n$ with RM model and the SFT model. Specifically, for each prompt in the dataset, we sample $n$ responses from the SFT model, and pick the response with highest reward based on the RM or ground truth reward (GT) as the final response. Our results are shown in Table~\ref{tab:my_label}.

The performance gap between RM and ground truth reward is small in training and in-distribution test set, and it is much larger in the OOD dataset, especially when $n$ is large. We also note that best-of-$n$ greatly boosts the performance of SFT model especially in the in-distribution test set (from $43.4\%$ to $60\%$).
Intuitively, best-of-$n$ accuracy can be treated as an upper bound of model performance after RL training with this RM.

\begin{figure}
    \centering
    \includegraphics[width=1\linewidth]{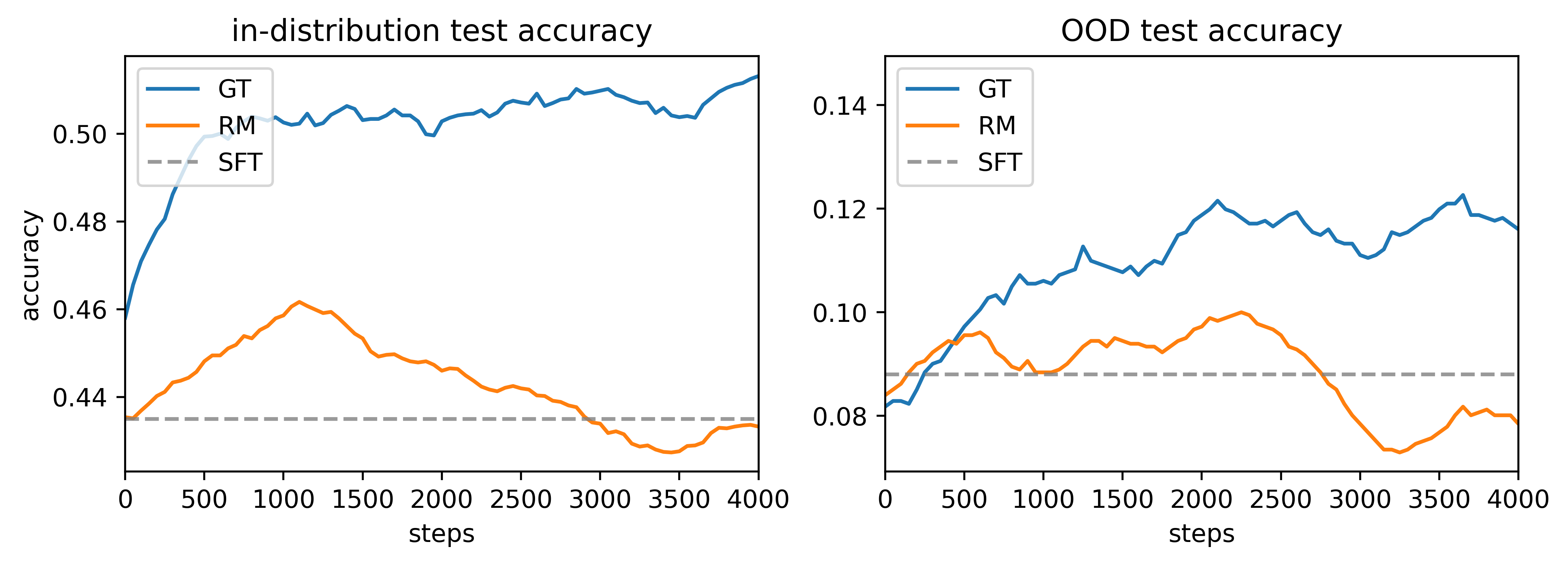}
    \caption{Performance of PPO with ground-truth reward and RM. While PPO with the ground truth reward keeps boosting both the in-distribution and OOD accuracies, PPO with RM start to see performance degradation after a short period of training.}
    \label{fig:ppo}
\end{figure}

\subsection{PPO}
In contrast to previous studies, we set $\beta=0$ to eliminate the KL regularization against the SFT model during PPO training. This is not a matter of concern when training with a ground-truth reward, as it consistently provides an accurate learning signal. However, during training with a RM, we found KL regularization to be ineffective: a strong KL regularization impedes the model's learning, while insufficient KL regularization leads to a drop in performance.
In response to this, we propose a novel type of regularization. The underlying concept is to constrain the policy to always remain within the top-$p$ vocabulary of the reference model. We defer the full implementation details of PPO with extra discussion to Appendix \ref{appendix:ppo}.




\noindent{\bf Results.}
Results are shown in Figure~\ref{fig:ppo}.
We see that in both in-distribution and OOD dataset, PPO with ground truth reward consistently improves model performance as the number of training steps increase.

PPO with RM enhances model performance in the initial 2,500 steps. However, post this phase, performance declines below baseline as the model begins to intentionally generate responses that were not encountered by the RM during its training, effectively ``hacking'' the RM. More detailed evidence of model ``hacking'' the RM is included in Appendix \ref{appendix:ppo}.

\subsection{DPO and IPO}

The first step is to construct the preference dataset \( \mathcal{D}_{\text{pref}} = \{(x^{(i)}, y_w^{(i)}, y_l^{(i)})\}_{i = 1}^N \). We begin with generating responses based on a given prompt \( x \) using the SFT model with a temperature of \( 1 \) to encourage diversity in the responses. Each generated response is then evaluated using the ground truth reward function \( r^*(x, y) \) to determine its quality. If a response is assigned a reward less than \( 1.0 \), indicating either a format discrepancy or incorrect reasoning, it is included in the preference dataset as a rejected response \( y_l \). The corresponding chosen response \( y_w \) is sourced from the SFT dataset, ensuring that \( y_w \) adheres to both the correct format and reasoning process. Through this method, we construct a preference dataset of size \( 86,000 \). We initialize the model from the SFT model and train it for 2 epochs with a batch size of 128. Throughout this phase, we employ a learning rate of \( 5 \times 10^{-7} \) with a constant scheduler.



\noindent{\bf Regularization of DPO and IPO.}
In our experimental analysis of DPO and IPO methodologies, a critical observation is regarding the regularization practices inherent to each approach. This inspires us to conduct a detailed investigation into the role of the regularization parameter \( \beta \) within DPO and IPO frameworks.

\begin{figure}
    \centering
    \includegraphics[width=1\linewidth]{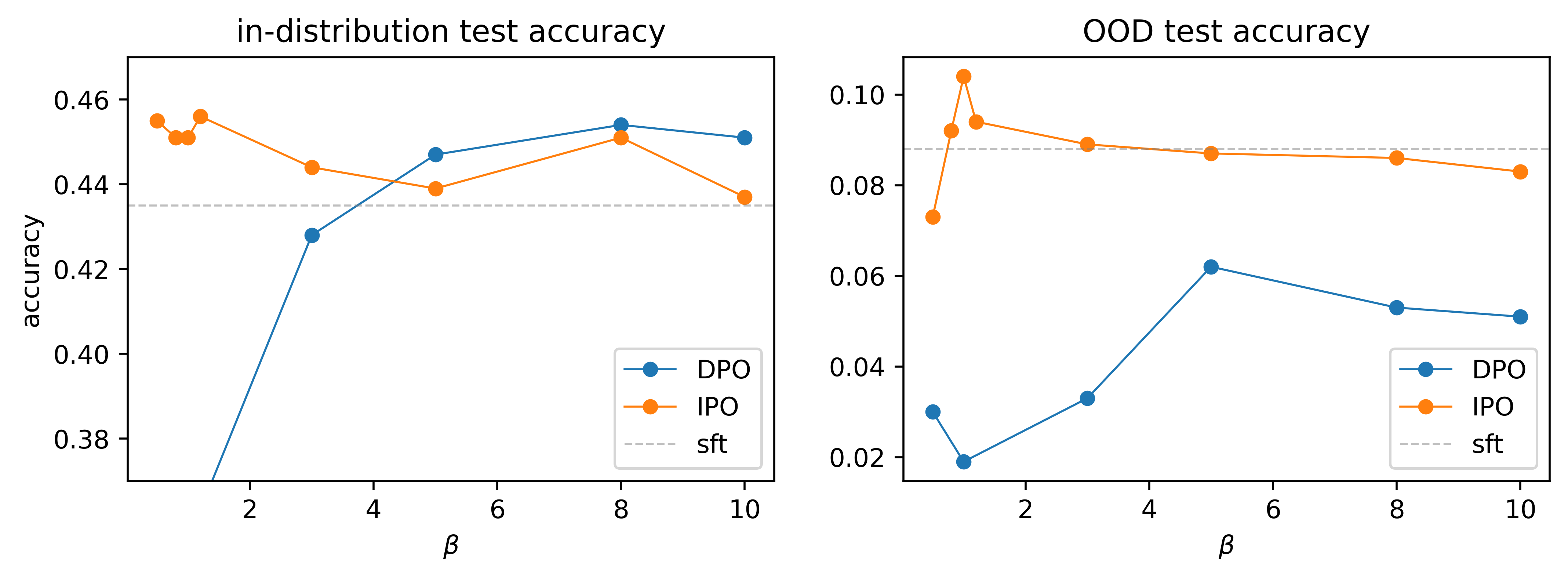}
    \caption{Test accuracy for DPO and IPO for different values of regularization parameter $\beta$. We observe that IPO is more robust across different $\beta$. It is worth noting that both DPO and IPO fail to enhance generalization of the generative LMs.}
    \label{fig:dpo_beta}
\end{figure}

The results of this investigation are shown in Figure \ref{fig:dpo_beta}. We see that a larger \( \beta \) value is requisite for DPO to attain stability and mitigate overfitting in the in-distribution test. However, even with an optimized \( \beta \), DPO's performance in OOD test set failed to surpass that of the baseline SFT model, showing the limitations of DPO's OOD generalization ability.
In comparison, IPO demonstrates better performance relative to SFT across both in-distribution and OOD test set, which is not sensitive to the \( \beta \) values explored. This resilience and performance edge of IPO can be attributed to its inherently stronger regularization framework, which effectively balances the model's fidelity to the training data with its capacity to generalize to the OOD setting.



\section{Conclusion}
Leveraging the $(N, K)$-Puzzle as a testbed revealed some valuable insights into RL training for generative LMs. While PPO with ground truth rewards consistently improved performance, PPO with a trained RM suffered late-stage performance drops due to the RM being ``hacked''. In contrast, although DPO and IPO avoid RM training for simplicity of implementation, we observe limited generalization from in-distribution to out-of-distribution prompts, constraining their potential in developing versatile generative LMs. We see significant potential for further exploration of the $(N, K)$-Puzzle as a testbed, which serves as a standardized and cost-effective environment for identifying the most effective strategies for RL training in generative LMs.

\section{Ethical Statement}
In this study, we explore the application of Reinforcement Learning (RL) within Large Language Models (LLMs) through the lens of a generalized version of the 24-Puzzle. In our opinion, this domain appears innocuous and devoid of direct ethical considerations due to its abstract and game-based nature.

\section{Limitations}
One of the primary limitations of the current study lies in the scale of the LMs employed. Our research utilizes models of a relatively modest size, which, while sufficient in the context of a generalized 24-Puzzle, may not fully capture the complexities and capabilities inherent to larger-scale LMs. The constrained model size can limit the depth and breadth of language understanding and generation, potentially impacting the generalizability of our findings to more sophisticated tasks and broader applications.


\bibliography{custom, anthology}

\clearpage
\appendix

\input{rm_ppo}\

\end{document}

%% file: rm_ppo.tex
\section{Examples of Responses with Different Ground Truth Reward} \label{sec:eg-reward}
Given the following prompt,
\begin{description}
    \item {\it Prompt:} 24;2,3,4,6:
\end{description}
we include examples of responses with different ground truth rewards as follows,

\begin{itemize}
    \item {\it Response with reward 1.0:} \\ \qquad 4+6=10,10-2=8,8*3=24 
    \item {\it Response with reward 0.1:} \\ \qquad 4+6=10,10-2=8,8+3=11
    \item {\it Response with reward 0.0:} \\ \qquad 4+6=10,10-2=8,8*3=22
\end{itemize}

\section{Detail and Discussion of PPO}\label{appendix:ppo}
In all of our PPO experiments, we use policy clip range $\epsilon_{\rm pg} = 0.15$, value function clip range $\epsilon_{\rm vf} = 10.0$, GAE parameter $\lambda = 1.0$, and value function parameter $0.1$. On top of the standard formulation in \eqref{eq:rl}, we employ entropy regularization on the LM policy $\pi_\theta$ with a weight of $0.04$ to encourage exploration.

We train the LM policy for 2.56M episodes (prompt-response pairs). When optimizing the clipped PPO objective, we use AdamW \citep{loshchilov2017decoupled} with minibatches of size $512$ each. The learning rate decays at the rate of $0.98$ every $10$ PPO epochs (or $100$ steps).

\subsection{Training Details of PPO with Ground Truth Reward}
We train the LM policy for 2.56M episodes (prompt-response pairs). We use $5120$ episodes per PPO batch, which is optimized using AdamW \citep{loshchilov2017decoupled} for one epoch with $10$ minibatches of size $512$ each.

\begin{table}[ht]
    \centering
    \begin{tabular}{c|c}
        \hline
        Hyperparameter & Value \\
        \hline
        learning rate & $1e-5$\\
        \hline
        learning rate decay ratio & $0.98$ \\
        \hline
        discount factor $\gamma$ & $0.99$\\
        \hline
        GAE $\lambda$ & $1.0$ \\
        \hline
        PPO clip ratio & $0.15$ \\
        \hline
        batch size & $512 \times 10$ \\
        \hline
        entropy loss coefficient & $0.04$\\
        \hline
        minibatch size & $512$ \\
        \hline
    \end{tabular}
    \caption{Hyperparameters of PPO with ground truth reward.}
    \label{tab:PPO.RM.hyperparameter}
\end{table}

\subsection{Training Details of PPO with RM}
The hyperparameters used for training PPO with a reward model is in Table~\ref{tab:PPO.RM.hyperparameter}.
We found that KL regularization is not very effective in our task: it is either too strong and stop the algorithm from learning, or it is too weak so that the performance drops.
To resolve this, we propose a weaker regularization: for each response generated by the current policy, we fit it into the SFT model and determine whether each token of the response is in SFT model's top-$p$ vocabulary given input before this token.
If any of the token is out of the top-$p$ vocabulary, we change the reward of current response to the value of OOD penalty.
Intuitively, such regularization ensures that the trained policy only generates responses within the support of nucleus sampling of the original SFT policy, but it does not force the distribution of the trained policy to be close to the SFT policy.
This ensures that the support of the learnt policy distribution does not leave the ``trust region'' of the reward model.
Moreover, it does not force the policy distribution to be close to that of reference model, which might impede learning.

\begin{table}[ht]
    \centering
    \begin{tabular}{c|c}
        \hline
        Hyperparameter & Value \\
        \hline
        learning rate & $1e-6$\\
        \hline
        learning rate decay ratio & $0.98$ \\
        \hline
        discount factor $\gamma$ & $0.93$\\
        \hline
        GAE $\lambda$ & $1.0$ \\
        \hline
        PPO clip ratio & $0.1$ \\
        \hline
        batch size & $512 \times 88$ \\
        \hline
        entropy loss coefficient & $0.04$\\
        \hline
        minibatch size & $512$ \\
        \hline
        top-$p$ regularization &  $0.95$\\
        \hline
        OOD penalty & $-40$ \\
        \hline
    \end{tabular}
    \caption{Hyperparameters of PPO with RM}
    \label{tab:PPO.RM.hyperparameter}
\end{table}

\subsection{Examples of Reward Model Hacking}
\begin{figure}[t!]
    \centering
    \includegraphics[width=1\linewidth]{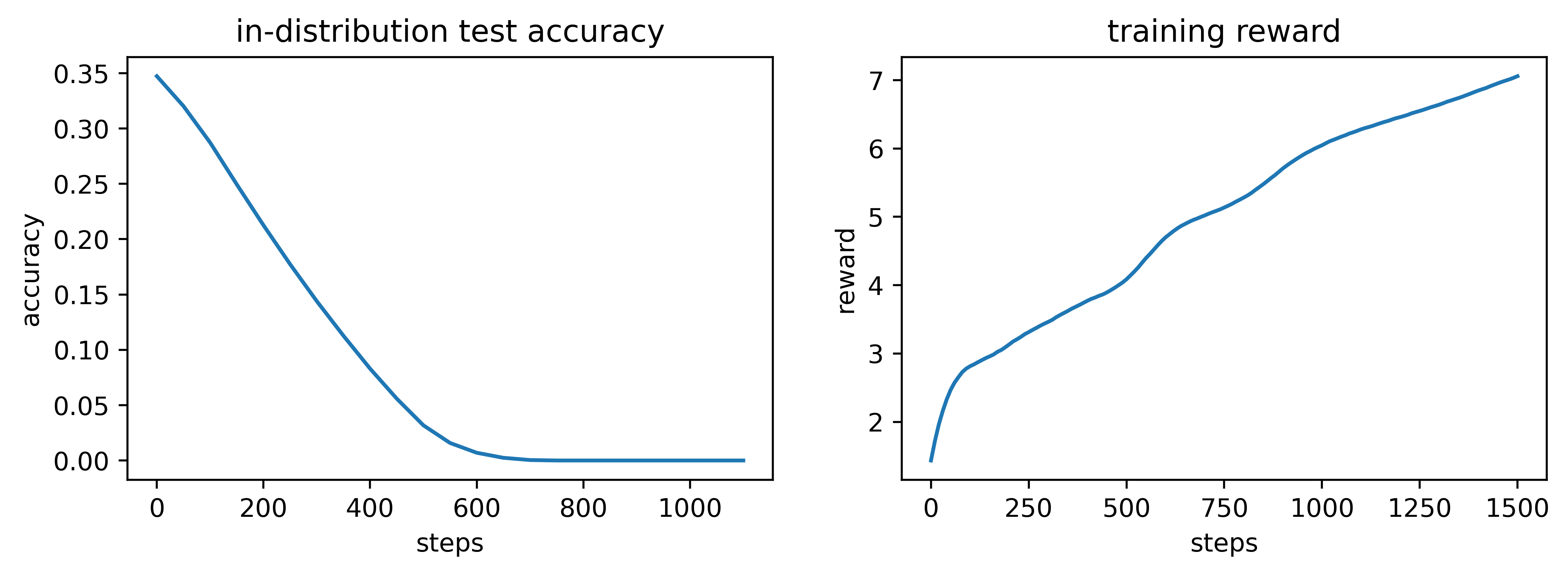}
    \caption{Training dynamics of PPO with reward model without KL or top-$p$ regularization.}
    \label{fig:ppo.beng}
\end{figure}
We also include samples generated from a policy trained by PPO without KL or top-$p$ regularization, whose performance quickly drops to zero as training proceeds (see Figure~\ref{fig:ppo.beng}).
On inspecting the samples, we find that the reward model is completely `hacked': many incorrect samples have very high rewards.
\begin{Verbatim}[commandchars=\\\{\}]
episode:\textcolor{cyan}{24;3,3,7,6:6*3=18,7-3=4,18+4=22}
reward model:\textcolor{cyan}{3.284614324569702}
ground truth reward: 0.0
episode:13;7,7,7,2:7-7=1,7*2=14,14-1=13
reward:8.917851448059082
ground truth reward: 1.0
episode:24;2,2,7,7:7-2=5,7+5=12,2*12=24
reward:6.0330657958984375
ground truth reward: 1.0
episode:\textcolor{cyan}{24;4,3,2,7:7-4=3,3+3=6,2*6=12}
reward:\textcolor{cyan}{3.4907703399658203}
ground truth reward: 0.0
episode:\textcolor{cyan}{24;3,7,4,2:3*7=21,4-2=2,21+2=23}
reward:\textcolor{cyan}{4.20366096496582}
ground truth reward: 0.0
episode:18;5,2,4,8:8/4=2,5*2=10,2*10=20
reward:1.4202626943588257
ground truth reward: 0.0
\end{Verbatim}

As evidence by the example rewards in \textcolor{cyan}{cyan} color, the incorrect responses can lead to moderate-to-high rewards, which are far less separated than the ground truth rewards. In the later stage of training, model tend to stick to some incorrect responses with moderate-to-high rewards.